\documentclass{article}

\usepackage{microtype}
\usepackage{graphicx}
\usepackage{subcaption}
\usepackage{booktabs} %
\usepackage{titlesec}
\usepackage{listings}
\usepackage{xcolor}
\usepackage{algorithm}
\usepackage{algpseudocode}
\usepackage{float}
\titlespacing*{\subsubsection}
{0pt}    %
{2pt}    %
{2pt}    %

\usepackage{enumitem}

\setlist[itemize]{
    topsep=2pt,
    itemsep=1pt,
    parsep=0pt,
    partopsep=0pt
}

\usepackage{hyperref}

\usepackage[accepted]{icml2026}

\usepackage{amsmath}
\usepackage{amssymb}
\usepackage{mathtools}
\usepackage{amsthm}
\usepackage{listings}

\usepackage[capitalize,noabbrev]{cleveref}

\theoremstyle{plain}

\theoremstyle{definition}

\theoremstyle{remark}

\usepackage[textsize=tiny]{todonotes}

\icmltitlerunning{QueryGraph: Reliable Multi-Tool Query Planning}

\begin{document}

\twocolumn[
  \icmltitle{QueryGraph: Reliable Multi-Tool Query Execution Planning via\\ LLM-Based Graph Generation}

  \icmlsetsymbol{equal}{*}

  \begin{icmlauthorlist}
    \icmlauthor{Aishwarya Chakravarthy}{equal,yyy}
    \icmlauthor{Vidhi Kulkarni}{equal,yyy}
    \icmlauthor{Duen Horng Chau}{yyy}
\end{icmlauthorlist}

\icmlaffiliation{yyy}{School of Computational Science and Engineering, Georgia Institute of Technology, Atlanta, GA, USA}

\icmlcorrespondingauthor{Aishwarya Chakravarthy}{achakrav6@gatech.edu}

  \icmlkeywords{Machine Learning, ICML}

  \vskip 0.3in
]

\printAffiliationsAndNotice{\icmlEqualContribution}  %

\begin{abstract}
Many real-world queries over personal data span multiple applications and require structured planning, as individual tools expose only partial information. While LLMs show strong reasoning and tool use, reliably executing multi-step, cross-tool queries remains challenging. We introduce a system that converts natural language queries into structured graphs and executes them via a deterministic planner. Our approach uses depth-first search to resolve dependencies and combine results across tools, improving reliability and enabling queries beyond traditional keyword-based search. We demonstrate high accuracy even with smaller or locally hosted LLMs.
\end{abstract}

\begin{figure*}[t]
    \centering
    \includegraphics[width=0.89\textwidth]{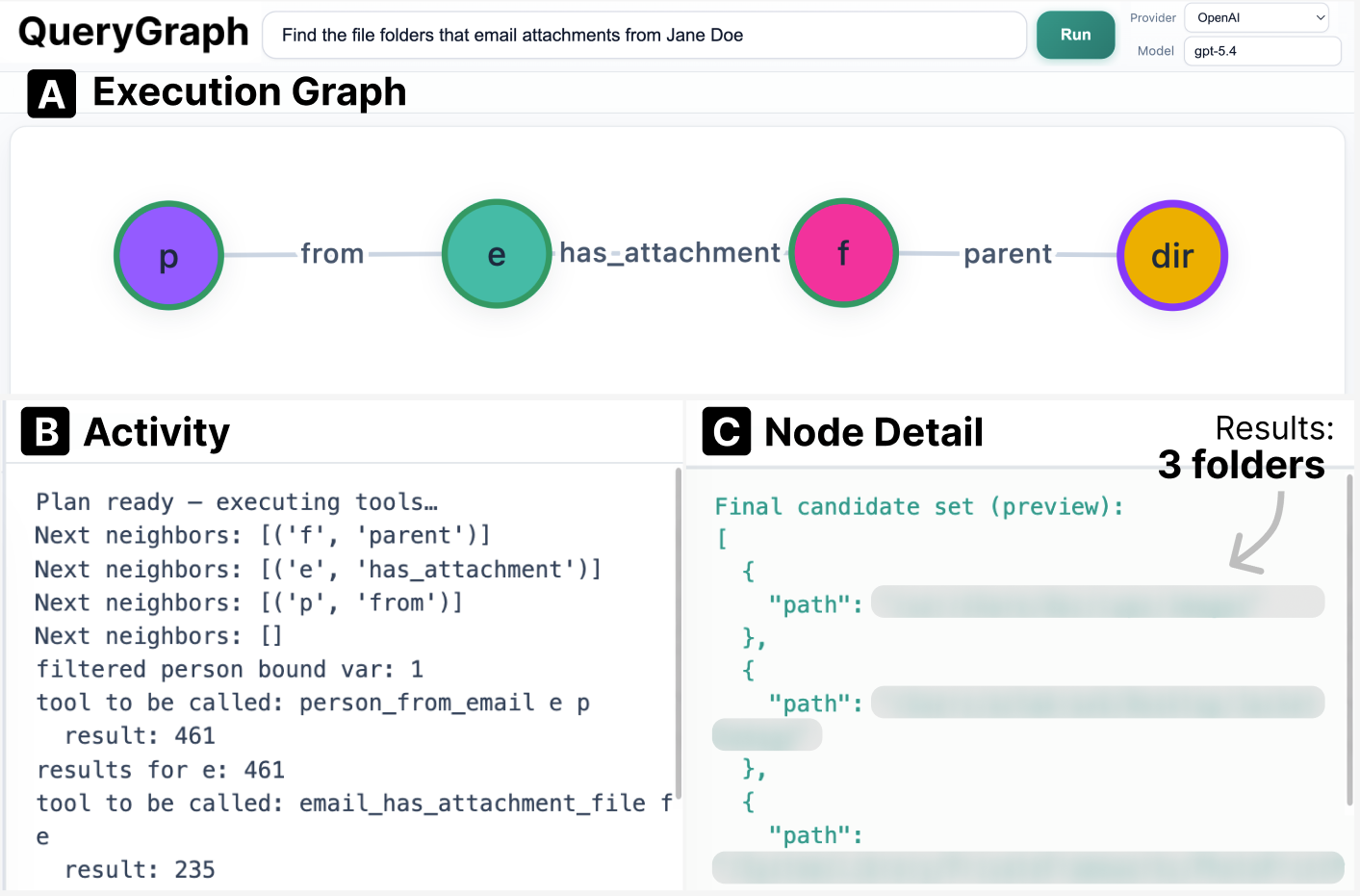}
    \caption{QueryGraph interface for \textit{``Find the file folders that contain email attachments from Jane Doe.''} 
    \textbf{A.} Execution graph with nodes for \textit{people} (p), \textit{emails} (e), \textit{files} (f), and \textit{directories} (dir), and edges such as \textit{from}, \textit{has\_attachment}, and \textit{parent}, showing stepwise resolution (p $\rightarrow$ e $\rightarrow$ f $\rightarrow$ dir). 
    \textbf{B.} Activity panel with intermediate results (e.g., 461 emails, 235 with attachments). 
    \textbf{C.} Node details showing the final candidate set.}
    \label{fig:figure_1}
\end{figure*}

\vspace{-20pt}
\section{Introduction}
While LLMs have shown promising capabilities in reasoning and tool use, recent work has also shown that their performance can degrade on complex, multi-step tasks, especially when planning and tool coordination are required\cite{aghzal2024path}. Executing queries that span multiple tool environments therefore remains a challenge. However, executing queries that contain multiple steps across different tool environments remains a challenge.

Such challenges are especially apparent in operating system environments like Windows and macOS, where applications (e.g., Calendar, Email) expose limited cross-application access. Existing tools such as Spotlight and Windows Search rely on keyword-based retrieval and support only simple queries. As a result, queries like \textit{``What is the name of the person that I met last week at Piedmont Park?"} require multiple coordinated steps: identifying dates, retrieving event attendees, and linking contact and location data, which current systems cannot execute end-to-end.

We introduce \textbf{QueryGraph}, a system for cross-application query resolution that builds on Feldspar’s retrieval framework \cite{feldspar} and extends it with LLM-based planning. Our approach converts natural language queries into structured graphs, where entities are nodes and relationships are edges, with variables optionally bounded by filters. Using this representation, QueryGraph defines a planning algorithm that determines tool call order by performing a depth-first search (DFS) from the return variable to resolve dependencies.

Our main contributions include:

\begin{enumerate}[itemsep=0pt, topsep=0pt, parsep=0pt, partopsep=0pt, leftmargin=4mm]
\item \textbf{A consistent tool-call planning framework for LLMs.} QueryGraph enables LLMs to resolve multi-step queries across applications via tool calls. It converts queries into graphs and uses an execution planner to improve reliability over traditional agentic tool calling.

\item \textbf{Strong Performance with Local LLMs.} QueryGraph demonstrates high planning accuracy even with smaller, locally hosted models. By translating queries into graphs and using an execution planner, our method reduces LLM risk and avoids reliance on large proprietary models.

\item \textbf{Improved Query Capability Beyond Traditional Search.} Our method outperforms traditional search systems like Spotlight, which rely on keyword similarity. By reasoning across multiple applications, it resolves multi-tool queries that conventional systems cannot answer.

\item \textbf{An Interactive User Interface for Query Execution and Visualization.} We develop an interactive UI that lets users input natural language queries and visualize both the graph representation and step-by-step node resolution (Figure~\ref{fig:figure_1}), improving transparency across tools.
\end{enumerate}

We have demonstrated the efficacy of our tool by performing testing across multiple tools, such as Mail, Calendar, and Files. Our results highlighted that structured query planning systems substantially improve the accuracy of multiple-step query execution.

\begin{figure*}[t]
\centering
\includegraphics[width=\textwidth]{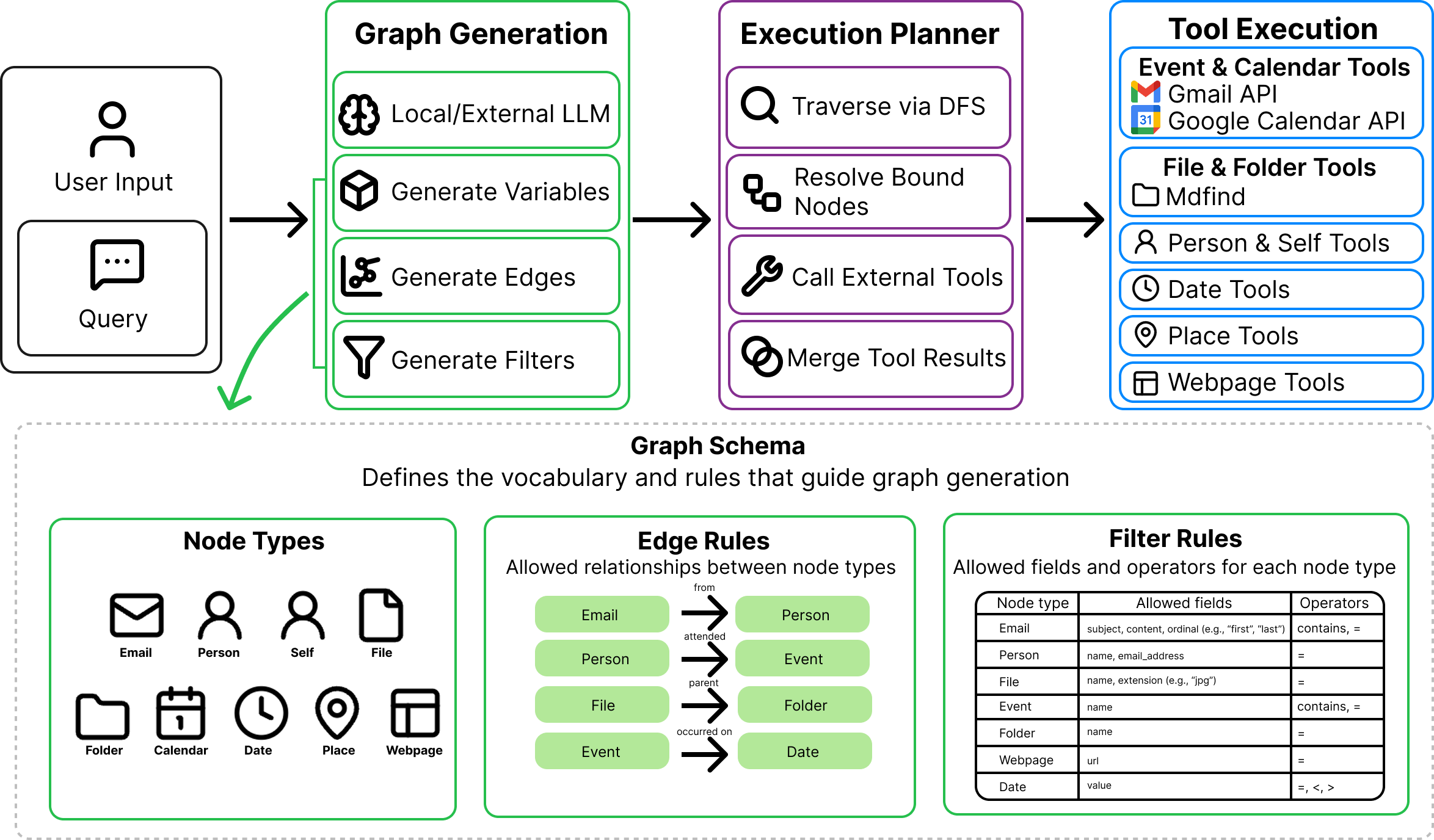}
\caption{QueryGraph architecture illustrating the transformation of natural language queries into graph representations and their execution via a structured planning and tool-calling pipeline.}
\label{fig:architecture}
\end{figure*}

\section{Related Work}

Recent work explores how large language models (LLMs) extend beyond text generation to structured reasoning and execution, with a key challenge being coordination between planning and tool use. ReAct \cite{yao2023react} and Toolformer \cite{schick2023toolformer} interleave reasoning with tool calls, while Plan-and-Solve prompting \cite{wang2023planandsolve} separates planning and execution to improve accuracy.

LLMs can also generate structured plans, such as object-level plans or transition functions \cite{paulius2024bootstrapping, shlomi2025transition}, or act as high-level planners alongside classical frameworks \cite{tantakoun2025survey, puerta2025roadmap}. However, as task complexity increases, purely language-based reasoning becomes less reliable, motivating hybrid approaches that combine reasoning with execution \cite{aghzal2024path}.

Rather than linear reasoning, graph-based approaches model plans as interconnected steps that branch and merge. Methods such as Graph of Thoughts and Graph Chain-of-Thought \cite{besta2024graphthoughts, jin2024graphcot}, along with Tree-of-Traversals and GTool \cite{markowitz2024tree, chen2025gtool}, use structured traversal to guide multi-step, multi-tool tasks.

Building on these ideas, our work represents natural language queries as dependency graphs resolved through structured traversal and tool execution.

\section{Query Representation and System Foundations}

To enable LLM-based planning, we define node types
$S = {\text{Email}, \text{Person}, \text{Self}, \text{File}, \text{Folder}, \text{Event}, \text{Date}, \text{Place}, \text{Webpage}}$,
where \textit{Self} represents the user issuing the query. QueryGraph supports two actions: \texttt{find} to return matching entities, and \texttt{count} to return the number of matches. Each node type supports predefined edges and filters, detailed in Appendix~\ref{app:graph_representation}.

\section{Algorithm and System Design}

We now describe the steps of our algorithm, illustrated using the example query 
\textit{``Find the webpage mentioned in the email from the person you met in an event in February.''}

\subsection{Step 1: Construct Graph Representation}

We model the LLM as a function that maps a natural language query to a structured graph representation. Let $q \in \mathcal{Q}$ denote an input query, and let $S$ be the set of node types defined previously. We define the LLM as a function
\[
\mathcal{Q} \rightarrow \mathcal{G},
\]
where $\mathcal{G}$ is the space of valid graph representations.

Given a query $q$, the LLM produces a graph
\[
G = (V, E, F, a, r, m),
\]
where $V$ is the set of typed vertices, $E \subseteq V \times \mathcal{R} \times V$ is the set of typed edges, $F$ is the set of filter constraints of the form $(v, f, \mathrm{op}, \mathrm{val})$, and $a$, $r$, and $m$ denote the action, return variable, and return mode, respectively.

The LLM is prompted with the specification of node types $S$, edge rules $\mathcal{R}$, and allowable filters, and is required to output a valid graph $G$. Edge rules are treated as undirected, since enforcing directed edges would require the LLM to determine execution order explicitly; instead, traversal order is handled by the DFS-based planner. The detailed prompt to create the graph is shown in the Appendix~\ref{app:graph_generation_prompt}.

\begin{figure}[t]
    \centering
    \includegraphics[width=\columnwidth]{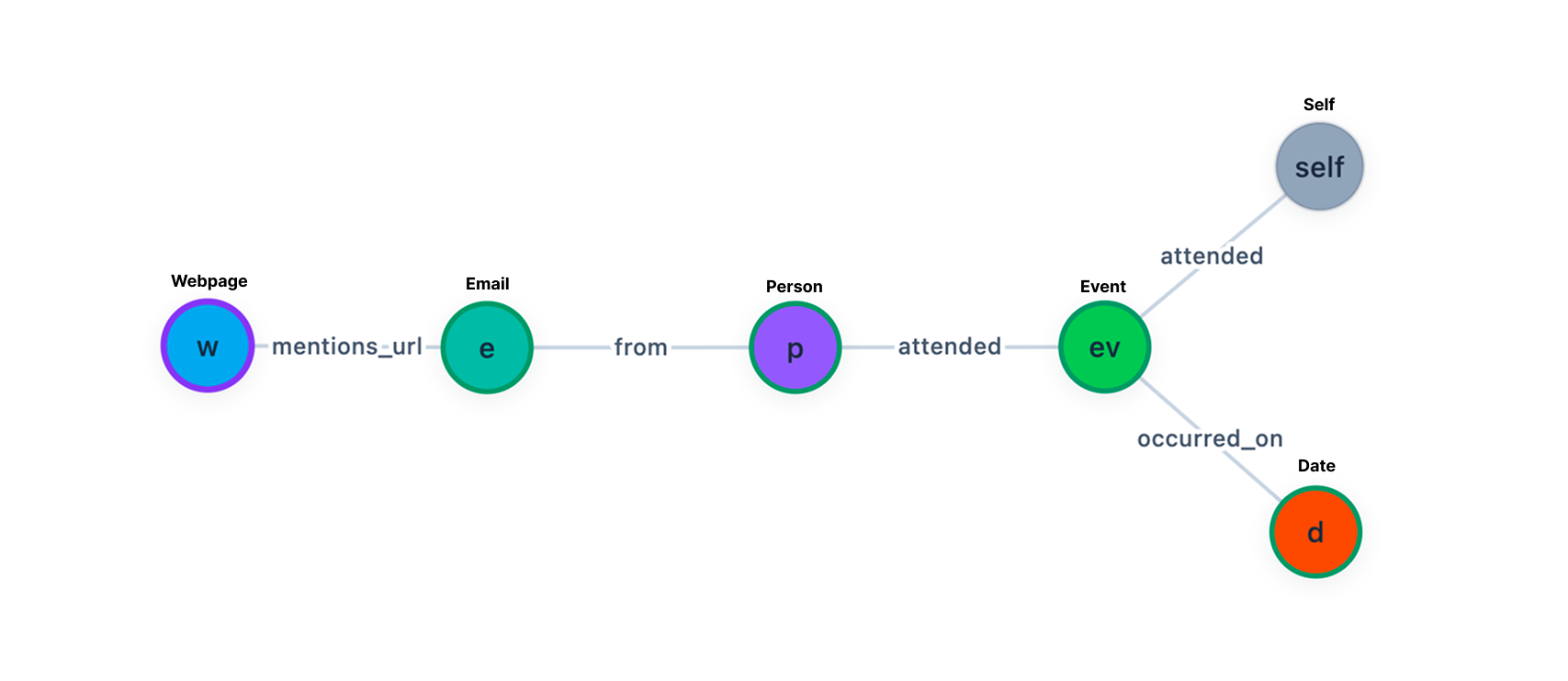}
    \caption{Graph representation of a sample query in the QueryGraph UI. Nodes denote entities: web results, emails, person, events, date, and the current user (self). Edges represent relations such as \textit{mentions\_url}, \textit{from}, \textit{attended}, and \textit{occurred\_on}.}
    \label{fig:figure_2}
\end{figure}

For the example query, the LLM produces the following graph:
{\small
\begin{lstlisting}[
    basicstyle=\tiny, %
]
Graph = {
  ``action'': ``find'',
  ``return_var'': ``w'',
  ``return_mode'': ``all'',
  ``vars'': {
    ``w'': ``Webpage'',
    ``e'': ``Email'',
    ``p'': ``Person'',
    ``self'': ``Self'',
    ``ev'': ``Event'',
    ``d'': ``Date''
  },
  ``constraints'': [
    {``kind'': ``edge'', ``from'': ``e'', ``edge'': ``mentions_url'', ``to'': ``w''},
    {``kind'': ``edge'', ``from'': ``e'', ``edge'': ``from'', ``to'': ``p''},
    {``kind'': ``edge'', ``from'': ``p'', ``edge'': ``attended'', ``to'': ``ev''},
    {``kind'': ``edge'', ``from'': ``self'', ``edge'': ``attended'', ``to'': ``ev''},
    {``kind'': ``edge'', ``from'': ``ev'', ``edge'': ``occurred_on'', ``to'': ``d''},
    {``kind'': ``filter'', ``var'': ``d'', ``field'': ``value'', ``op'': ``='', ``value'': ``in February''}
  ]
}
\end{lstlisting}
}
\vspace{-10pt}
We then postprocess this output by converting the edge and filter constraints into an adjacency list representation, treating edges as undirected for traversal purposes. The visualization of the graph can be seen in Figure~\ref{fig:figure_2}.

\subsection{Step 2: Depth-First Search Execution}

We execute queries using depth-first search (DFS) over the graph, maintaining a visited set $V$ and a mapping $C$ from variables to candidate sets. The traversal runs in \(O(|V| + |E|)\). Starting from the return variable, DFS proceeds recursively. Nodes with no unvisited neighbors are treated as \emph{bound nodes}, whose candidate sets are determined solely by local filters. After resolving a bound node, the algorithm backtracks, propagating and combining candidate sets via edges and filters. Each edge supports bidirectional tools (left-to-right and right-to-left), enabling resolution in either direction (e.g., people $\leftrightarrow$ events). See Appendix~\ref{app:algorithm_pseudocode} for pseudocode.
\vspace{-5pt}
\subsection{Evaluation of Sample Prompt}

For the example query, the return variable is the webpage. After performing DFS, the algorithm identifies two bound variables, \textit{date} and \textit{self}, which can be resolved independently of their neighbors. The \textit{self} node is trivially resolved, while \textit{date} requires normalization due to potentially vague expressions (e.g., \textit{``a few months ago,''} \textit{``last year''}). We use an LLM to map such expressions to a concrete date range (Appendix~\ref{app:date_normalization_prompt}).

\begin{figure}[t]
    \centering
    \includegraphics[width=\columnwidth]{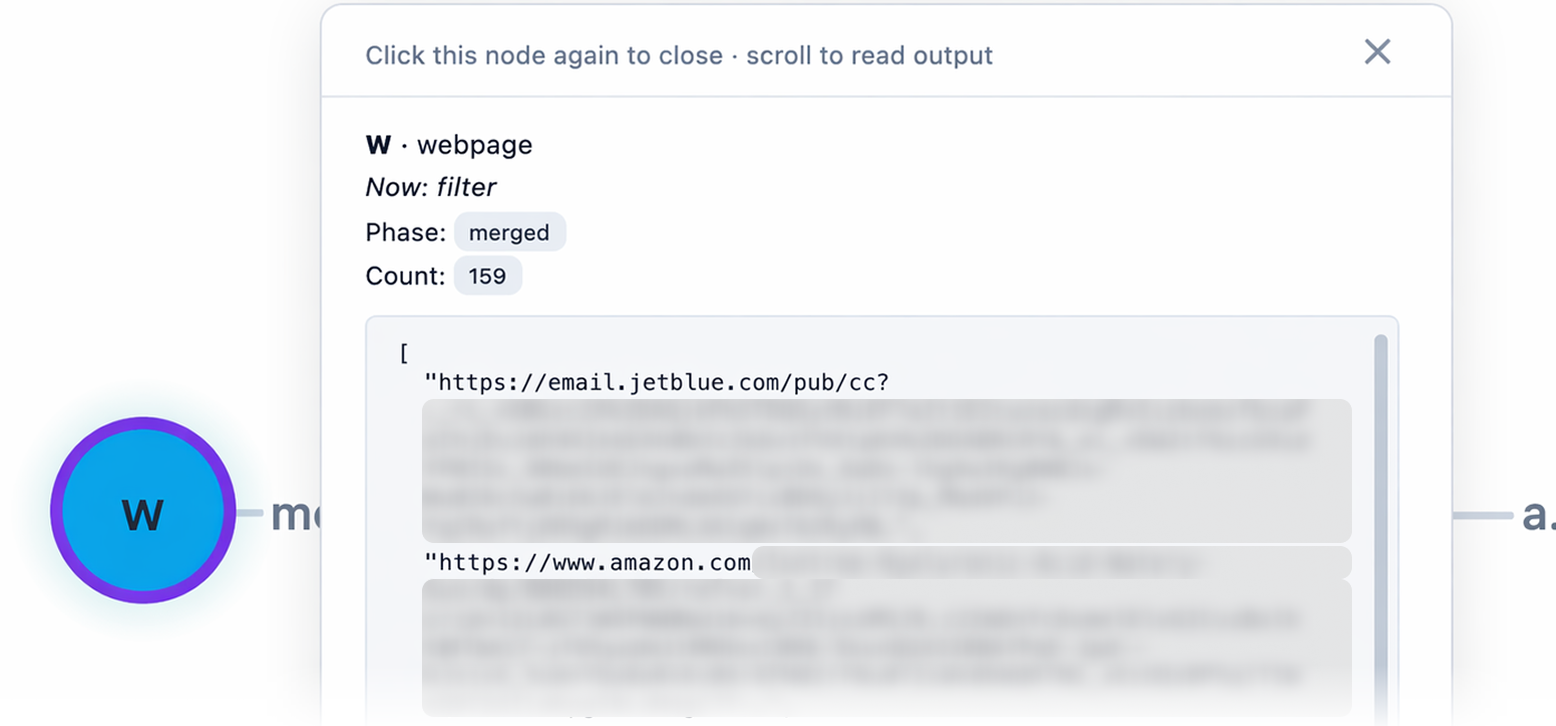}
    \caption{Final output of the sample query in the QueryGraph UI. The figure shows the resolved results for the node \textit{w} (webpage) after executing the full query plan. The system returns a filtered set of webpages (159 results), displaying concrete outputs (e.g., links from JetBlue emails, Amazon product pages).}
    \label{fig:figure_3}
\end{figure}

Once resolved, these variables are used to compute candidate events via tool calling (intersecting results from the \emph{self\_attended\_event} left to right tool and \emph{event\_occurred\_on\_date} right to left tool). The algorithm then propagates upward through the graph, resolving \emph{person}, \emph{email}, and finally the target \emph{webpage}. The final result is shown in Figure~\ref{fig:figure_3}.

\vspace{-10pt}

\vspace{3pt}
\subsection{Architecture Design}
QueryGraph connects each entity type to its underlying data source. Remote data is handled via Google Workspace APIs: Gmail supports email search, content, and attachments, while Calendar provides access to events and metadata (e.g., time, location, attendees). The system supports both local and remote LLMs with local tool execution, reducing external data transmission. On macOS, file and folder retrieval uses Spotlight via \texttt{mdfind}, enabling efficient name- and extension-based queries over the indexed filesystem. The web interface is a lightweight Flask application that streams execution updates to the frontend, with the execution graph dynamically rendered using D3.js.

\section{Results}

In order to measure the efficacy of QueryGraph, we evaluated the accuracy of the tool on queries designed to be difficult to search for on traditional operating systems, based on queries introduced in Feldspar~\cite{feldspar}. We tested the tool using both an external and local LLM.

The evaluated queries are:
\begin{enumerate}[itemsep=0mm, topsep=0mm, parsep=0mm, partopsep=0mm, leftmargin=3mm]
    \item Find the last email received on January 10, 2025.
    
    \item Find all the email attachments of type \texttt{.txt}.
    
    \item Find out who had email conversations with the person who sent out the file \texttt{file name}.
    
    \item Find out who attended the event in which \texttt{person name} was present.
    
    \item Find all the events that were attended by anyone who has sent you a file.
    
    \item Find the file folders that contain email attachments from \texttt{person name}.
    
    \item Find the webpage mentioned in the email from the person you met at an event in February.
\end{enumerate}

We log the graph generation accuracy, execution accuracy, and average runtime when utilizing GPT-5.4 (external model) and GPT-OSS:20B with medium thinking (local model) in Table~\ref{tab:results}. The graph visualization and logs for each query are placed in the Appendix~\ref{app:results}. 

\begin{table}[t]
\centering
\small
\setlength{\tabcolsep}{4pt}
\begin{tabular}{lccc}
\toprule
\textbf{Model} & \textbf{Graph (\%)} & \textbf{Exec (\%)} & \textbf{Avg. Time (s)} \\
\midrule
GPT-5.4 & 100 & 100 &  30.469\\
GPT OSS (medium) & 100 & 100 & 61.161 \\
\bottomrule
\end{tabular}
\caption{Graph generation accuracy, execution accuracy, and average runtime across models.}
\label{tab:results}
\end{table}

\section{Future Work}
In the future, we plan to reduce the rigidity of our graph schema to incorporate additional node types, edges, and filters based on user studies of real-world queries. Additionally, we plan to investigate adaptive traversal strategies for query execution, including DFS approaches that prioritize the most restrictive constraints to reduce intermediate search space and improve efficiency.

\bibliographystyle{icml2026} 
\bibliography{references}   

\clearpage
\appendix

\section*{Appendix}
\addcontentsline{toc}{section}{Appendix}

\renewcommand{\thesection}{A\arabic{section}}

\section{Prompt Construction}
\subsection{Graph Generation Prompt}
\label{app:graph_generation_prompt}
\begin{lstlisting}
You are a semantic query parser.

Your task is to convert a natural language query about personal data into a structured JSON representation describing:

- variables (typed entities)
- constraints between them
- filters on properties
- which variable should be returned

---

# Allowed entity types

Use these types when creating variables:

- Email
- Person
- Self
- File
- Event
- Folder
- Webpage
- Date
- Place

The entity Self represents the user issuing the query.

---

# JSON schema

Return JSON with the following fields:

- "action": the user intent
  Allowed values: "find", "count"

- "return_var": the variable name whose objects should be returned

- "return_mode": "one" or "all"
Default to "all". Use "one" only if the user's wording forces a single answer (superlative, ordinal, or unambiguous singular picking one entity). If it could be one or many, use "all".

- "vars": dictionary mapping variable names to entity types

- "constraints": list of constraints

Each constraint must be one of these:

## Edge constraint

Represents a relationship between two variables.

{
  "kind": "edge",
  "from": "<var>",
  "edge": "<relation>",
  "to": "<var>"
}

## Filter constraint

Represents a property condition on a variable.

{
  "kind": "filter",
  "var": "<var>",
  "field": "<field>",
  "op": "<operator>",
  "value": <value>
}

---

# Rules for Relations

Only use the relations listed below.

- Email "from" Person
- Email "to" Person
- Email "participants" Person
- Email "from" Self
- Email "to" Self
- Email "participants" Self
- Email "has_attachment" File
- Email "mentions_url" Webpage
- Email "sent_on" Date
- Person "attended" Event
- Self "attended" Event
- File "parent" Folder
- Event "occurred_on" Date
- Event "location" Place

---

## Rules for Filters 

### Allowed fields and operators per entity type

#### Email
- "subject"  
  operators: "contains"

- "content"  
  operators: "contains"

- "ordinal" (e.g., "last", "first")  
  operators: "="

---

#### Person
- "name"  
  operators: "=" 

- "email_address"  
  operators: "=" 
  
---

#### File
- "name"  
  operators: "=" 

- "extension"  
  operators: "=" (e.g. "extension" = "jpg")
  Do not include the period in the extension.

---

#### Event
- "name"  
  operators: "=", "contains"

---

#### Folder
- "name"  
  operators: "=" 

---

#### Webpage
- "url"  
  operators: "=" 

---

#### Date
- "value"
  operators: "=", ">", "<"

Use "=" for specific dates or bounded ranges (including relative ranges)
(e.g., "on Feb 2, 2026", "in April 2026", "from May to July 2025", "last week", "a couple of months ago")

Use "<" for strictly before a date
(e.g., "before July 2025", "prior to 01/15/2026")

Use ">" for on or after / since a date
(e.g., "after Monday", "since Feb 2026", "from 2020 onward")

---

#### Place
- "name"  
  operators: "=" 

### Rules

- Only use fields listed above for each entity type.
- The operator must be compatible with the field type.
- Do not invent new fields or operators.

---

# Variable naming

Use short lowercase variable names:

- e for Email
- p for Person
- self for Self
- f for File
- ev for Event
- dir for Folder
- w for Webpage
- d for Date
- pl for Place

If multiple are needed, use numbers (p1, p2, e1, e2).

---

# Important rules

- You may ONLY use the edges and filters allowed in the JSON schema.
- If the query implies a meeting, model it as an Event.
- If the query refers to a date expression like "July 27, 2007" or "a couple of months ago", represent it using a Date variable plus a filter.
- For Date filters, preserve the date expression exactly as it appears in the query.
- If the user refers to themself ("I", "me", "my"), use the Self entity.
- If the query implies that the user is involved in an Email or Event, you MUST include an edge connecting that variable to Self.
- Output JSON only. Do not include explanations or execution steps.

---

# Examples
...
\end{lstlisting}

\subsection{Date Normalization Prompt}
\label{app:date_normalization_prompt}
\begin{lstlisting}
You are a date normalization system.
Convert any date expression into a standardized string format.

Use the following rules:

1. FULL DATE  
If the input specifies a full date (day, month, year), output:
MM/DD/YYYY  
Example: "February 2, 2026" -> "02/02/2026"

2. MONTH + YEAR (or just month/year reference)  
If the input specifies a month (with or without a year), output:
MM/YYYY  
Example: "Feb 2026" -> "02/2026"  
Example: "May" -> "05"

3. RELATIVE DATES  
If the input is relative (e.g., "last month", "past 3 months", "yesterday"),  
use {today_date} as the reference point and compute the corresponding range.

Output format for ranges:
(start,end)

Where:
- start and end follow MM/YYYY or MM/DD/YYYY when appropriate
- start is the earliest date
- end is the latest date

Examples:
If today's date for the example is 02/15/2026, then:
- "last month" -> (01/2026, 02/2026)  
- "past 3 months" -> (12/2025, 02/2026)

4. GENERAL RULES
- Always zero-pad months and days (e.g., 02, not 2)
- Always use numeric formats (no text months in output)
- Never output ambiguous text (like "recently")
- If a range is implied, ALWAYS return a tuple (start, end)

Return only the normalized string. Do not include explanations.
\end{lstlisting}

\section{Graph Representation Details}
\label{app:graph_representation}
\subsection{Edge Rules}

We define a fixed set of allowable edge types between node types.

\subsubsection{Email}
\begin{itemize}
    \item \texttt{from} $\rightarrow$ \textbf{Person}, \textbf{Self}
    \item \texttt{to} $\rightarrow$ \textbf{Person}, \textbf{Self}
    \item \texttt{participants} $\rightarrow$ \textbf{Person}, \textbf{Self}
    \item \texttt{has\_attachment} $\rightarrow$ \textbf{File}
    \item \texttt{mentions\_url} $\rightarrow$ \textbf{Webpage}
    \item \texttt{sent\_on} $\rightarrow$ \textbf{Date}
\end{itemize}

\subsubsection{Person}
\begin{itemize}
    \item \texttt{attended} $\rightarrow$ \textbf{Event}
\end{itemize}

\subsubsection{Self}
\begin{itemize}
    \item \texttt{attended} $\rightarrow$ \textbf{Event}
\end{itemize}

\subsubsection{File}
\begin{itemize}
    \item \texttt{parent} $\rightarrow$ \textbf{Folder}
\end{itemize}

\subsubsection{Event}
\begin{itemize}
    \item \texttt{occurred\_on} $\rightarrow$ \textbf{Date}
    \item \texttt{location} $\rightarrow$ \textbf{Place}
\end{itemize}

\subsection{Filter Rules}

We define a set of allowable filter fields and operators for each node type.

\subsubsection{Email}
\begin{itemize}
    \item \textbf{subject}: supports the operator \texttt{contains}
    \item \textbf{content}: supports the operator \texttt{contains}
    \item \textbf{ordinal} (e.g., ``first'', ``last''): supports the operator \texttt{=}
\end{itemize}

\subsubsection{Person}
\begin{itemize}
    \item \textbf{name}: supports the operator \texttt{=}
    \item \textbf{email\_address}: supports the operator \texttt{=}
\end{itemize}

\subsubsection{File}
\begin{itemize}
    \item \textbf{name}: supports the operator \texttt{=}
    \item \textbf{extension}: supports the operator \texttt{=} (e.g., \texttt{extension = "jpg"})
\end{itemize}

\subsubsection{Event}
\begin{itemize}
    \item \textbf{name}: supports the operators \texttt{=} and \texttt{contains}
\end{itemize}

\subsubsection{Folder}
\begin{itemize}
    \item \textbf{name}: supports the operator \texttt{=}
\end{itemize}

\subsubsection{Webpage}
\begin{itemize}
    \item \textbf{url}: supports the operator \texttt{=}
\end{itemize}

\subsubsection{Date}
\begin{itemize}
    \item \textbf{value}: supports the operators \texttt{=}, \texttt{>}, and \texttt{<}
\end{itemize}

\section{Algorithm Pseudocode}
\label{app:algorithm_pseudocode}

\begin{algorithm}[H]
\caption{ExecutePlan}
\begin{algorithmic}[1]

\Require Graph $G$
\Ensure Result set or $\texttt{None}$

\State $nodes \gets G.nodes$
\State $r \gets G.return\_var$

\If{$r \notin nodes$}
    \State \Return $\texttt{None}$
\EndIf

\State $state.candidate\_sets \gets \emptyset$
\State $visited \gets \emptyset$

\Function{DFS}{$var, parent$}
    \State add $var$ to $visited$
    \State $neighbors \gets$ unvisited neighbors of $var$

    \If{$neighbors = \emptyset$}
        \State $res \gets \textsc{ResolveBoundVar}(var, parent)$
        \State $state.candidate\_sets[var] \gets res$
        \State \Return true
    \EndIf

    \State $curr \gets \texttt{null}$

    \For{each $(nbr, edge)$ in $neighbors$}
        \If{\Call{DFS}{$nbr, var$} = false}
            \State \Return false
        \EndIf

        \State $t \gets \textsc{RunTool}(nbr, edge, var, state)$

        \If{$t = \texttt{None}$}
            \State \Return false
        \EndIf

        \If{$curr = \texttt{null}$}
            \State $curr \gets t$
        \Else
            \State $curr \gets curr \cap t$
        \EndIf
    \EndFor

    \State $curr \gets \textsc{ApplyFilters}(var, curr)$
    \State $state.candidate\_sets[var] \gets curr$
    \State \Return true
\EndFunction

\State \Call{DFS}{$r, \texttt{None}$}

\State $results \gets state.candidate\_sets[r]$

\If{$results = \emptyset$}
    \State \Return $\texttt{None}$
\EndIf

\State \Return \textsc{DoAction}$(G.action, results, G.return\_mode)$

\end{algorithmic}
\end{algorithm}

\section{Results}
\label{app:results}
\subsection{GPT Results}
\begin{enumerate}
    \item Find the last email received on January 10, 2025. 
    \textbf{Graph Construction} \\
    \includegraphics[width=\columnwidth]{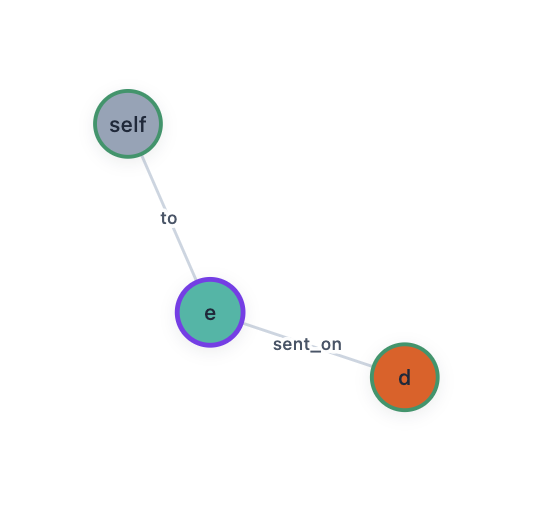}
    \textbf{Tool Execution Log} \\
    \begin{lstlisting}
    Next neighbors: [('self', 'to'), ('d', 'sent_on')]
    Next neighbors: []
    filtered self bound var size: 1
    tool to be called: self_to_email e self
      result size: 24842
    Next neighbors: []
    filtered date bound var size: 1
    tool to be called: date_sent_on_email e d
      result size: 17
    performing intersection
    result size for e: 1
    Total query time (plan + tool run): 12.64s 
    \end{lstlisting}
    
    \item Find all the email attachments of type \texttt{.txt}. \\
    \textbf{Graph Construction} \\
    \includegraphics[width=\columnwidth]{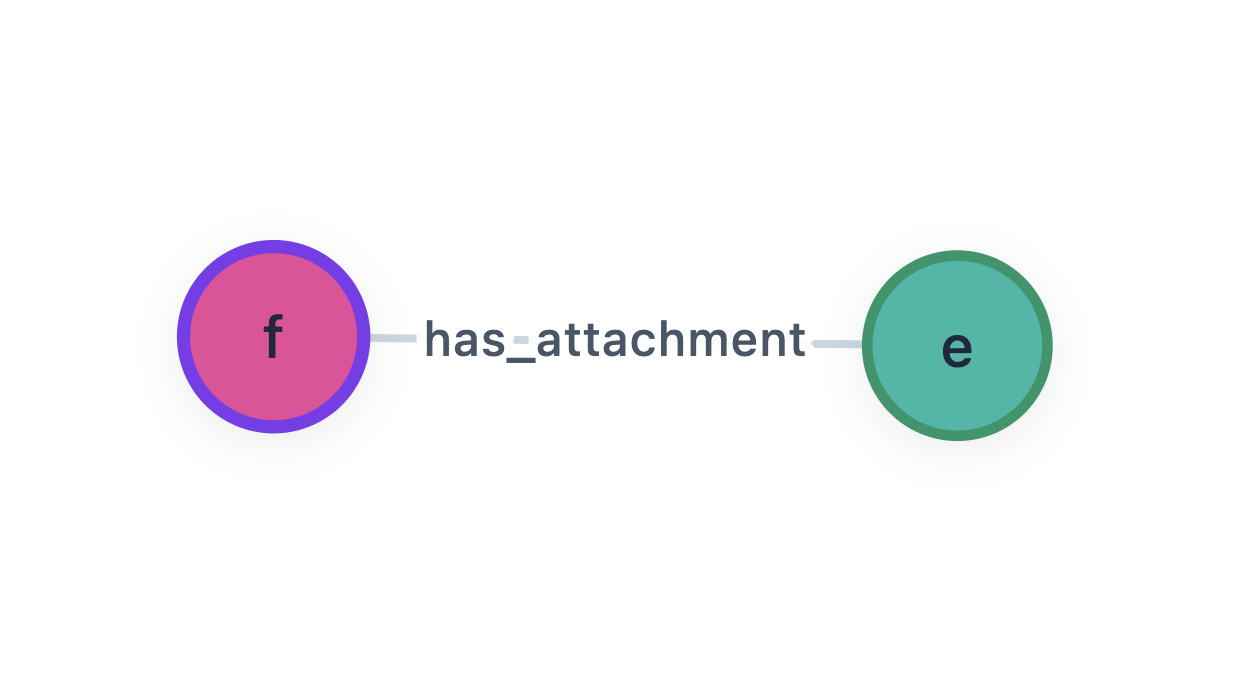}
    \textbf{Tool Execution Log} \\
    \begin{lstlisting}
    Next neighbors: [('e', 'has_attachment')]
    Next neighbors: []
    filtered email bound var: 'all'
    tool to be called: email_has_attachment_file f e
      result size: 986
    result size for f: 3
    Total query time (plan + tool run): 31.94s
    \end{lstlisting}
    
    \item Find out who had email conversations with the person who sent out the file \texttt{file name}. \\
    \textbf{Graph Construction} \\
    \includegraphics[width=\columnwidth]{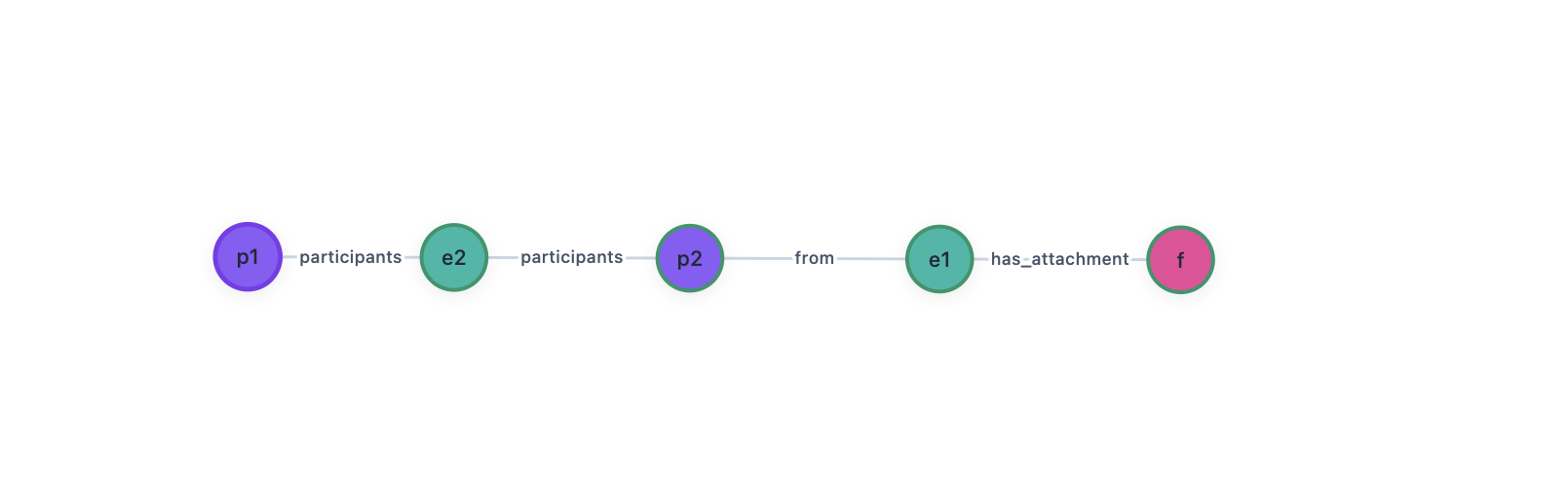}
    \textbf{Tool Execution Log} \\
    \begin{lstlisting}
    Next neighbors: [('e2', 'participants')]
    Next neighbors: [('p2', 'participants')]
    Next neighbors: [('e1', 'from')]
    Next neighbors: [('f', 'has_attachment')]
    Next neighbors: []
    filtered file bound var size: 1
    tool to be called: file_has_attachment_email e1 f
      result size: 1
    result size for e1: 1
    tool to be called: email_from_person p2 e1
      result size: 1
    result size for p2: 1
    tool to be called: person_participants_email e2 p2
      result size: 9
    result size for e2: 9
    tool to be called: email_participants_person p1 e2
      result size: 3
    result size for p1: 3
    Total query time (plan + tool run): 4.07s 
    \end{lstlisting}
    
    \item Find out who attended the event in which \texttt{person name} was present. \\
    \textbf{Graph Construction} \\
    \includegraphics[width=\columnwidth]{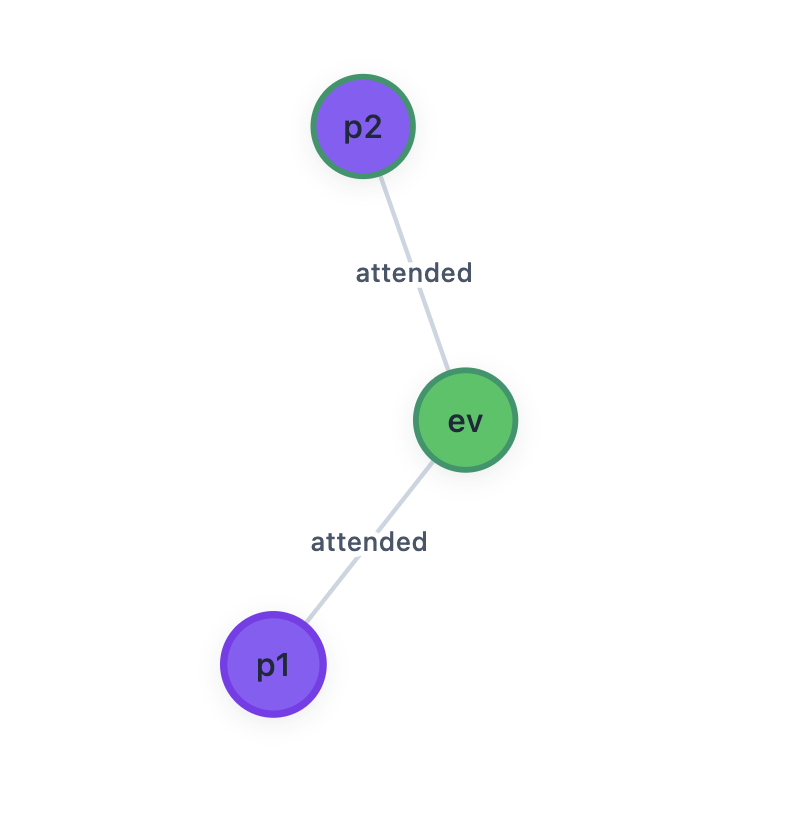}
    \textbf{Tool Execution Log} \\
     \begin{lstlisting}
    Next neighbors: [('ev', 'attended')]
    Next neighbors: [('p2', 'attended')]
    Next neighbors: []
    filtered person bound var size: 1
    tool to be called: person_attended_event ev p2
      result size: 2
    result size for ev: 2
    tool to be called: event_attended_person p1 ev
      result size: 1
    result size for p1: 1
    Total query time (plan + tool run): 3.64s
    \end{lstlisting}
    
    \item Find all the events that were attended by anyone who has sent you a file. \\
    \textbf{Graph Construction} \\
    \includegraphics[width=\columnwidth]{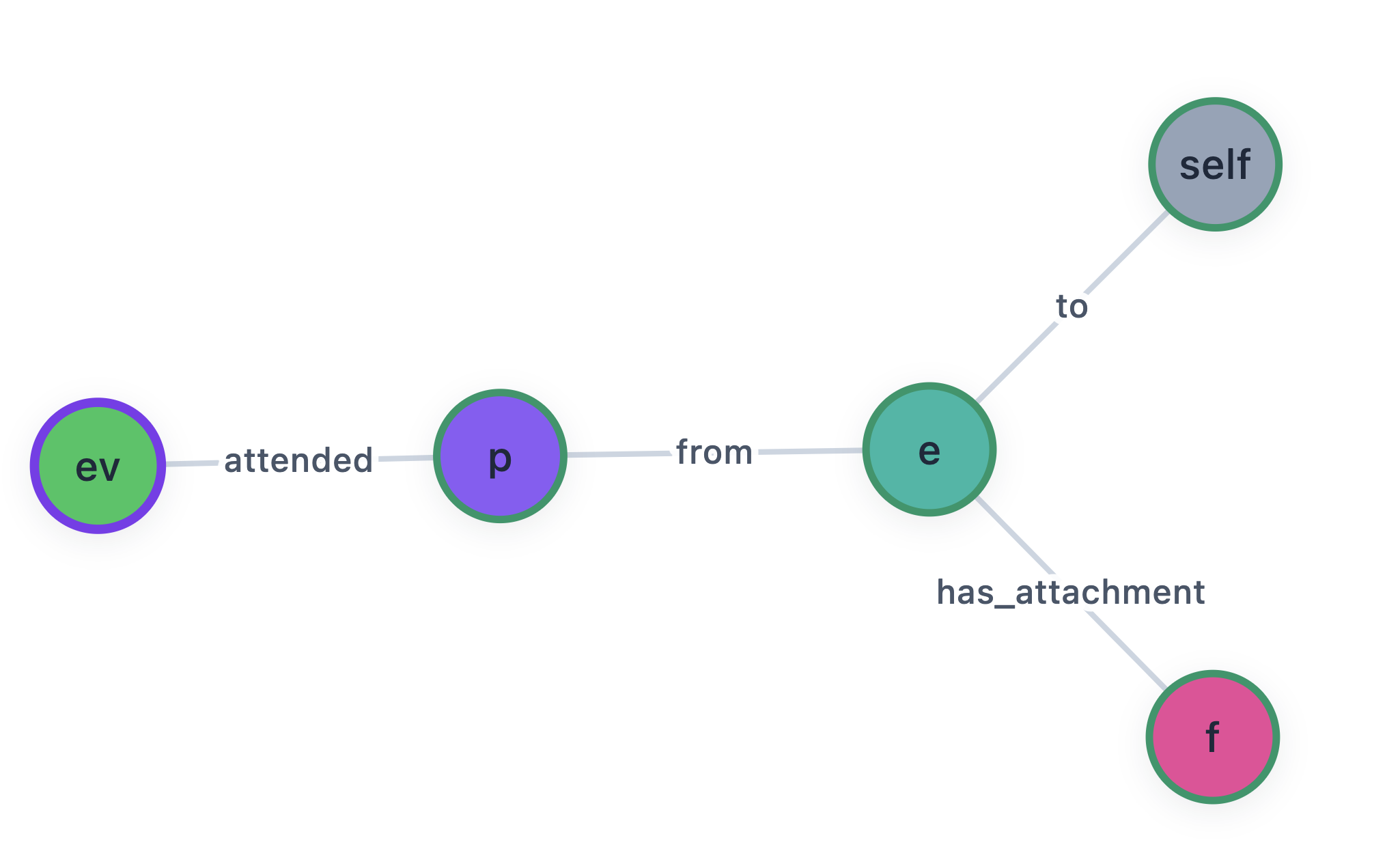}
    \textbf{Tool Execution Log} \\
    \begin{lstlisting}
    Next neighbors: [('p', 'attended')]
    Next neighbors: [('e', 'from')]
    Next neighbors: [('self', 'to'), ('f', 'has_attachment')]
    Next neighbors: []
    filtered self bound var size: 1
    tool to be called: self_to_email e self
      result size: 24842
    Next neighbors: []
    filtered file bound var size: 3
    tool to be called: file_has_attachment_email e f
      result size: 525
    performing intersection
    result size for e: 350
    tool to be called: email_from_person p e
      result size: 167
    result size for p: 167
    tool to be called: person_attended_event ev p
      result size: 81
    result size for ev: 81
    Total query time (plan + tool run): 101.54s 
    \end{lstlisting}
    \item Find the file folders that contain email attachments from \texttt{person name}. \\
    \textbf{Graph Construction} \\
    \includegraphics[width=\columnwidth]{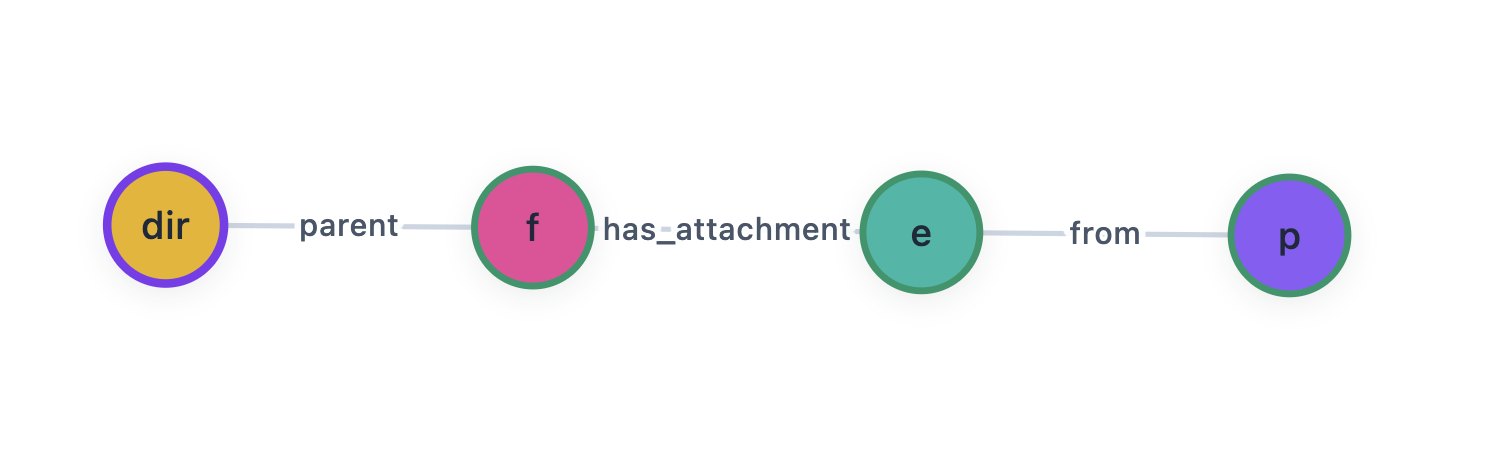}
    \textbf{Tool Execution Log} \\
     \begin{lstlisting}
    Next neighbors: [('f', 'parent')]
    Next neighbors: [('e', 'has_attachment')]
    Next neighbors: [('p', 'from')]
    Next neighbors: []
    filtered person bound var: 1
    tool to be called: person_from_email e p
      result size: 460
    result size for e: 460
    tool to be called: email_has_attachment_file f e
      result size: 235
    result size for f: 235
    tool to be called: file_parent_folder dir f
      result size: 6
    result size for dir: 6
    Total query time (plan + tool run): 44.73s
    \end{lstlisting}
    
    \item Find the webpage mentioned in the email from the person you met at an event in February. \\
    \textbf{Graph Construction} \\
    \includegraphics[width=\columnwidth]{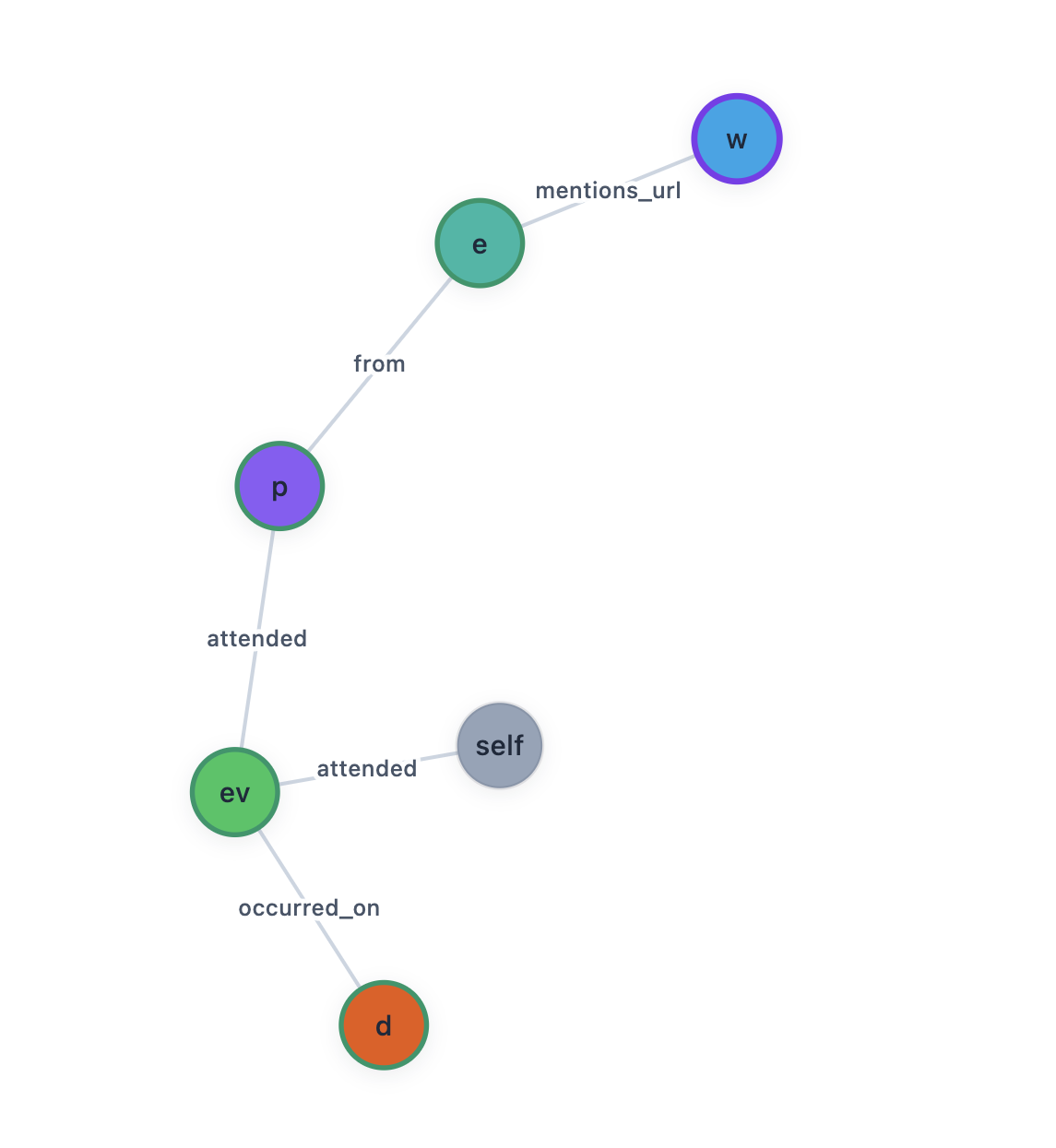}
    \textbf{Tool Execution Log} \\
    \begin{lstlisting}
    Next neighbors: [('e', 'mentions_url')]
    Next neighbors: [('p', 'from')]
    Next neighbors: [('ev', 'attended')]
    Next neighbors: [('d', 'occurred_on')]
    Next neighbors: []
    filtered date bound var size: 1
    tool to be called: date_occurred_on_event ev d
      result size: 57
    result size for ev: 57
    tool to be called: event_attended_person p ev
      result size: 4
    result size for p: 4
    tool to be called: person_from_email e p
      result size: 54
    result size for e: 54
    tool to be called: email_mentions_url_webpage w e
      result size: 159
    result size for w: 159
    Total query time (plan + tool run): 14.72s
    \end{lstlisting}
\end{enumerate}

\subsection{Gpt-oss:20b Results with Thinking = Medium}
\begin{enumerate}
    \item Find the last email received on January 10, 2025. \\
    \textbf{Graph Construction} \\
    \includegraphics[width=\columnwidth]{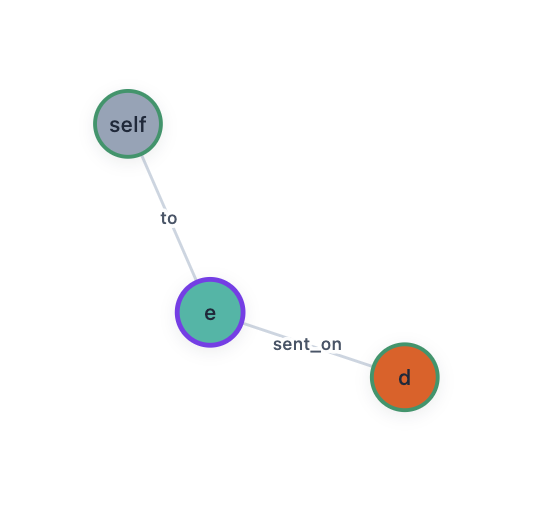}
    \textbf{Tool Execution Log} \\
    \begin{lstlisting}
    Next neighbors: [('self', 'to'), ('d', 'sent_on')]
    Next neighbors: []
    filtered self bound var size: 1
    tool to be called: self_to_email e self
      result: 24842
    Next neighbors: []
    filtered date bound var size: 1
    tool to be called: date_sent_on_email e d
      result size: 17
    performing intersection
    result size for e: 1
    Total query time (plan + tool run): 34.36s
    \end{lstlisting}
    
    \item Find all the email attachments of type \texttt{.txt}. \\
    \textbf{Graph Construction} \\
    \includegraphics[width=\columnwidth]{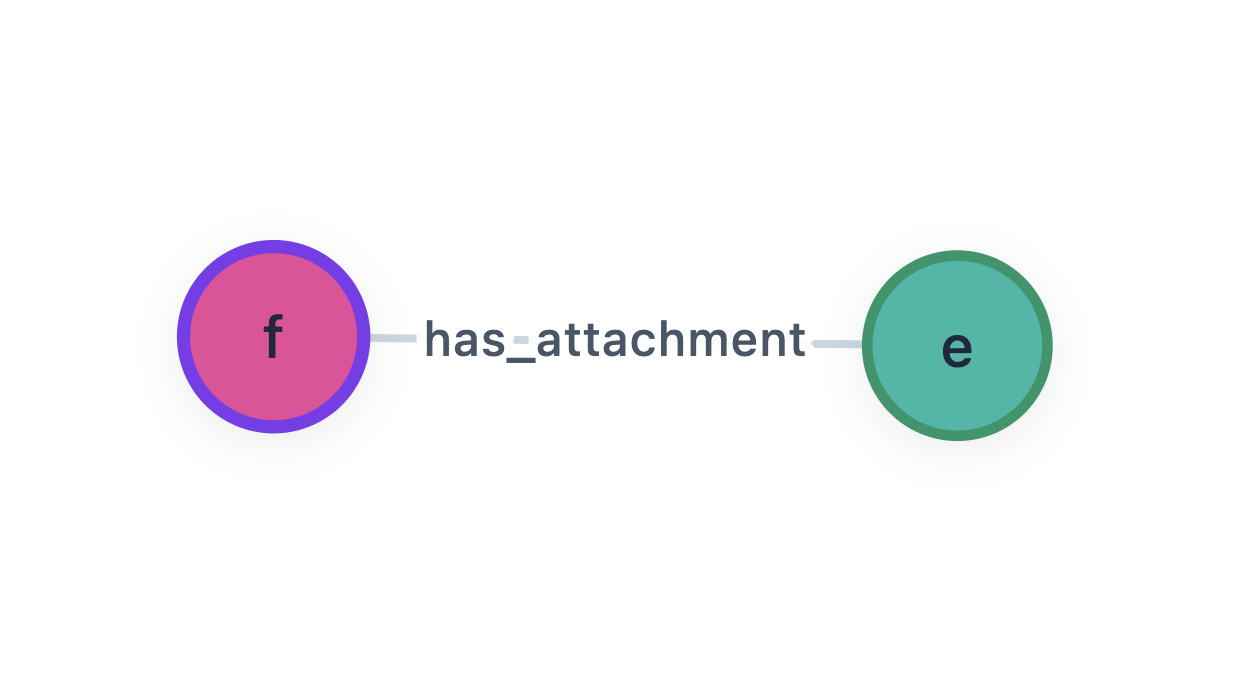}
    \textbf{Tool Execution Log} \\
     \begin{lstlisting}
    Next neighbors: [('e', 'has_attachment')]
    Next neighbors: []
    filtered email bound var: 'all'
    tool to be called: email_has_attachment_file f e
      result size: 986
    result size for f: 3
    Total query time (plan + tool run): 42.32s
    \end{lstlisting}
    
    \item Find out who had email conversations with the person who sent out the file \texttt{file name}. \\
    \textbf{Graph Construction} \\
    \includegraphics[width=\columnwidth]{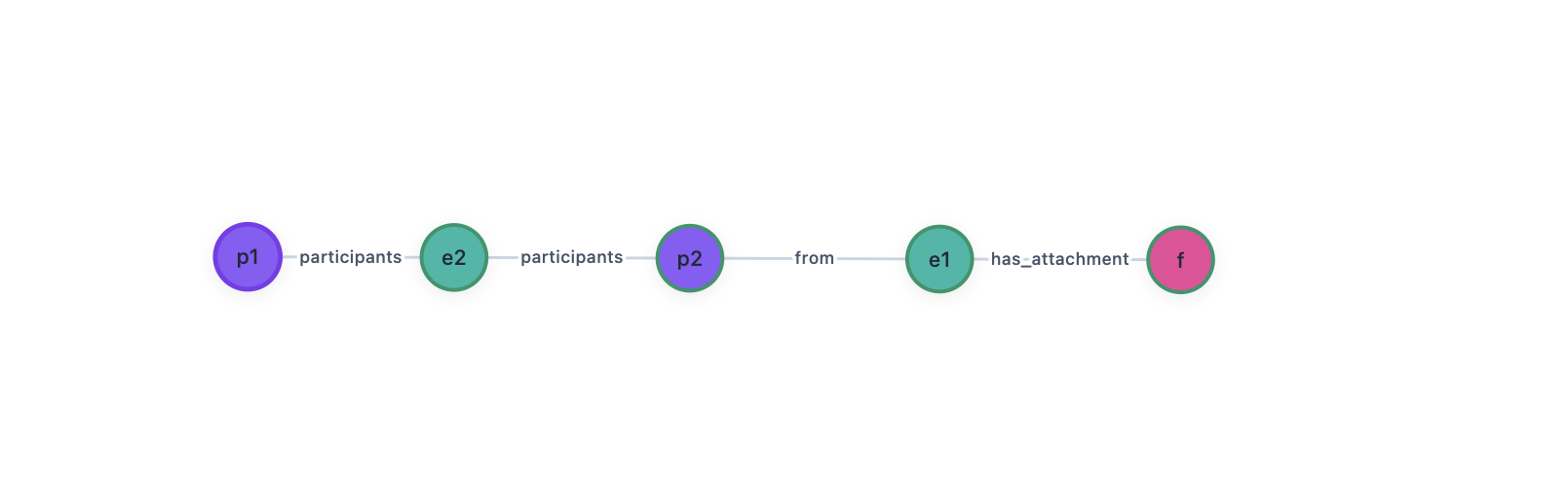}
    \textbf{Tool Execution Log} \\
    \begin{lstlisting}
    Next neighbors: [('e1', 'participants')]
    Next neighbors: [('p2', 'participants')]
    Next neighbors: [('e', 'from')]
    Next neighbors: [('f', 'has_attachment')]
    Next neighbors: []
    filtered file bound var: 1
    tool to be called: file_has_attachment_email e f
      result size: 1
    result size for e: 1
    tool to be called: email_from_person p2 e
      result size: 1
    result size for p2: 1
    tool to be called: person_participants_email e1 p2
      result size: 9
    result size for e1: 9
    tool to be called: email_participants_person p1 e1
      result size: 3
    result size for p1: 3
    Total query time (plan + tool run): 82.07s
    \end{lstlisting}
    
    \item Find out who attended the event in which \texttt{person name} was present. \\
    \textbf{Graph Construction} \\
    \includegraphics[width=\columnwidth]{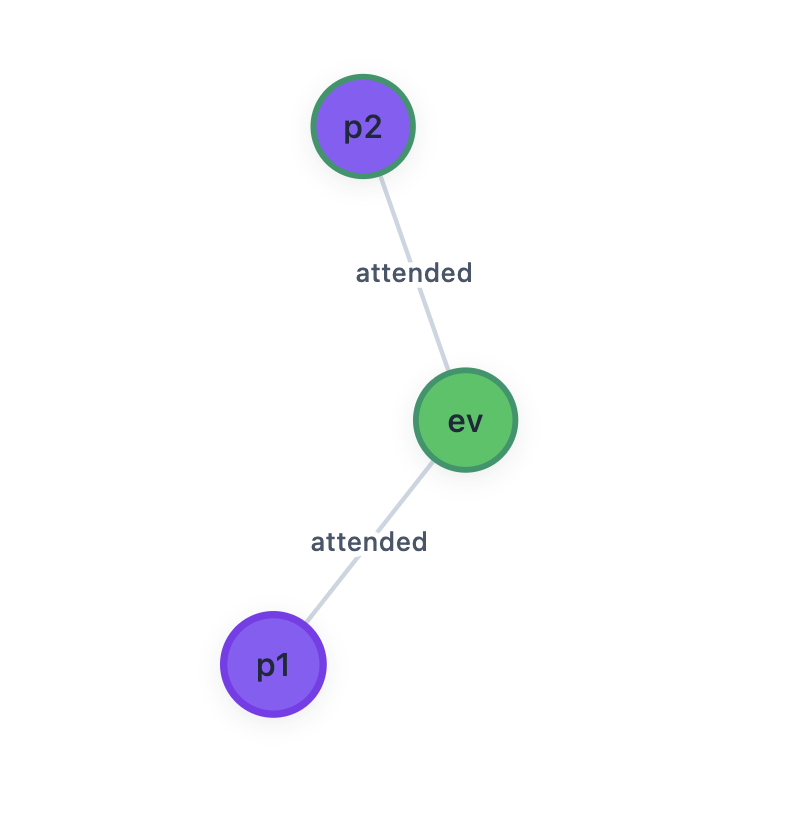}
    \textbf{Tool Execution Log} \\
    \begin{lstlisting}
    Next neighbors: [('ev', 'attended')]
    Next neighbors: [('p2', 'attended')]
    Next neighbors: []
    filtered person bound var size: 1
    tool to be called: person_attended_event ev p2
      result size: 2
    result size for ev: 2
    tool to be called: event_attended_person p1 ev
      result size: 1
    result size for p1: 1
    Total query time (plan + tool run): 18.68s
    \end{lstlisting}
    
    \item Find all the events that were attended by anyone who has sent you a file. \\
    \textbf{Graph Construction} \\
     \includegraphics[width=\columnwidth]{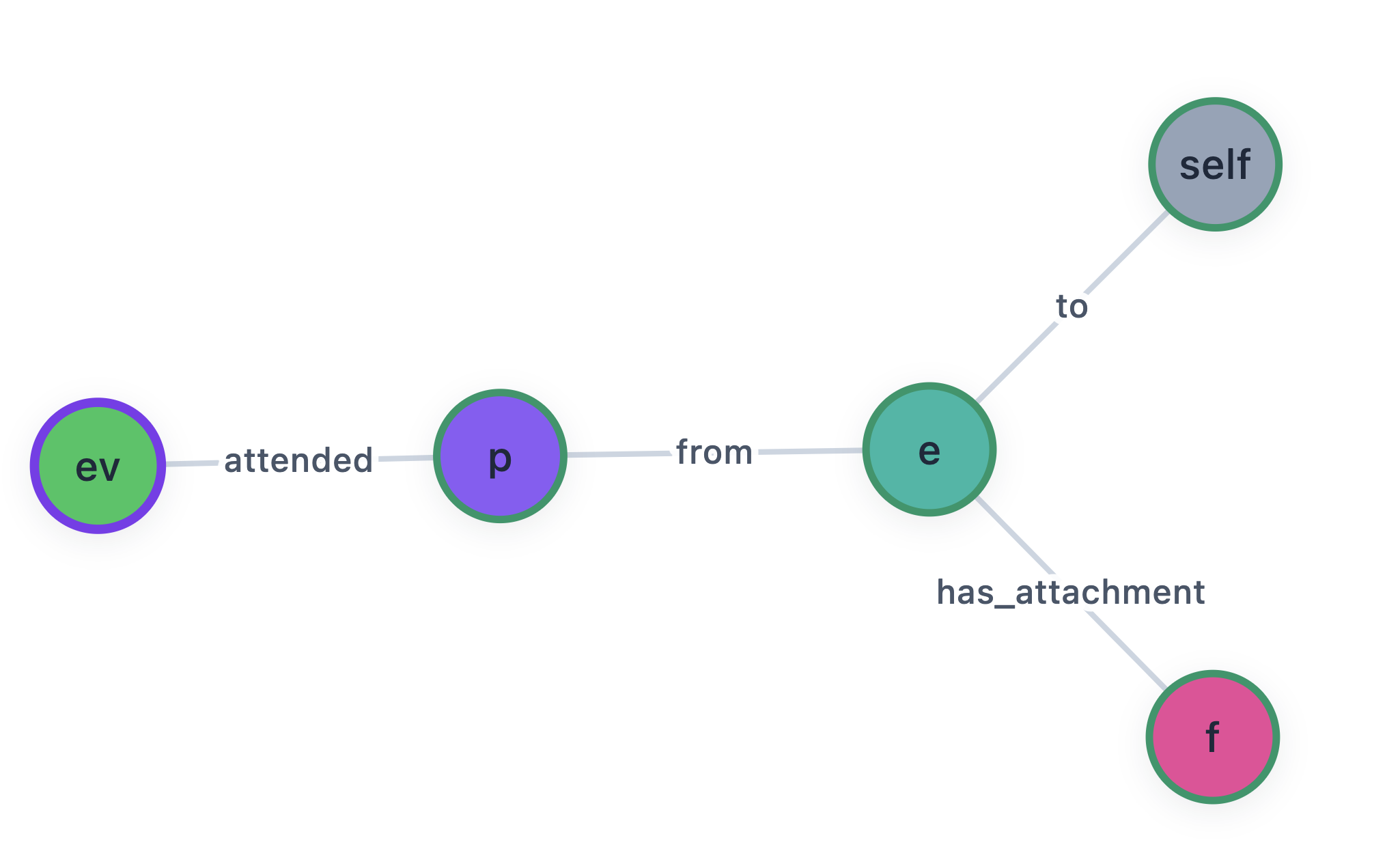}
    \textbf{Tool Execution Log} \\
    \begin{lstlisting}
    Next neighbors: [('p', 'attended')]
    Next neighbors: [('e', 'from')]
    Next neighbors: [('f', 'has_attachment'), ('self', 'to')]
    Next neighbors: []
    filtered file bound var size: 3
    tool to be called: file_has_attachment_email e f
      result size: 525
    Next neighbors: []
    filtered self bound var size: 1
    tool to be called: self_to_email e self
      result size: 24842
    performing intersection
    results for e size: 350
    tool to be called: email_from_person p e
      result size: 167
    result size for p: 167
    tool to be called: person_attended_event ev p
      result size: 81
    result size for ev: 81
    Total query time (plan + tool run): 126.22s
    \end{lstlisting}
    
    \item Find the file folders that contain email attachments from \texttt{person name}. \\
    \textbf{Graph Construction} \\
    \includegraphics[width=\columnwidth]{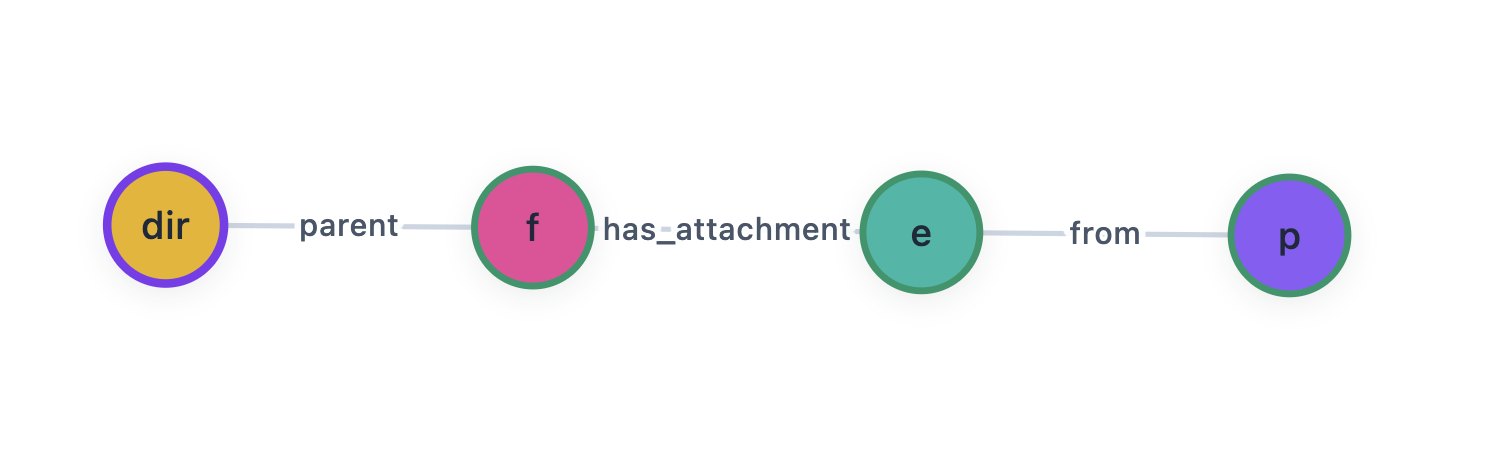}
    \textbf{Tool Execution Log} \\
    \begin{lstlisting}
    Next neighbors: [('f', 'parent')]
    Next neighbors: [('e', 'has_attachment')]
    Next neighbors: [('p', 'from')]
    Next neighbors: []
    filtered person bound var size: 1
    tool to be called: person_from_email e p
      result size: 460
    result size for e: 460
    tool to be called: email_has_attachment_file f e
      result size: 235
    result size for f: 235
    tool to be called: file_parent_folder dir f
      result size: 6
    result size for dir: 6
    Total query time (plan + tool run): 77.52s 
    \end{lstlisting}
    \item Find the webpage mentioned in the email from the person you met at an event in February. \\
    \textbf{Graph Construction} \\
    \includegraphics[width=\columnwidth]{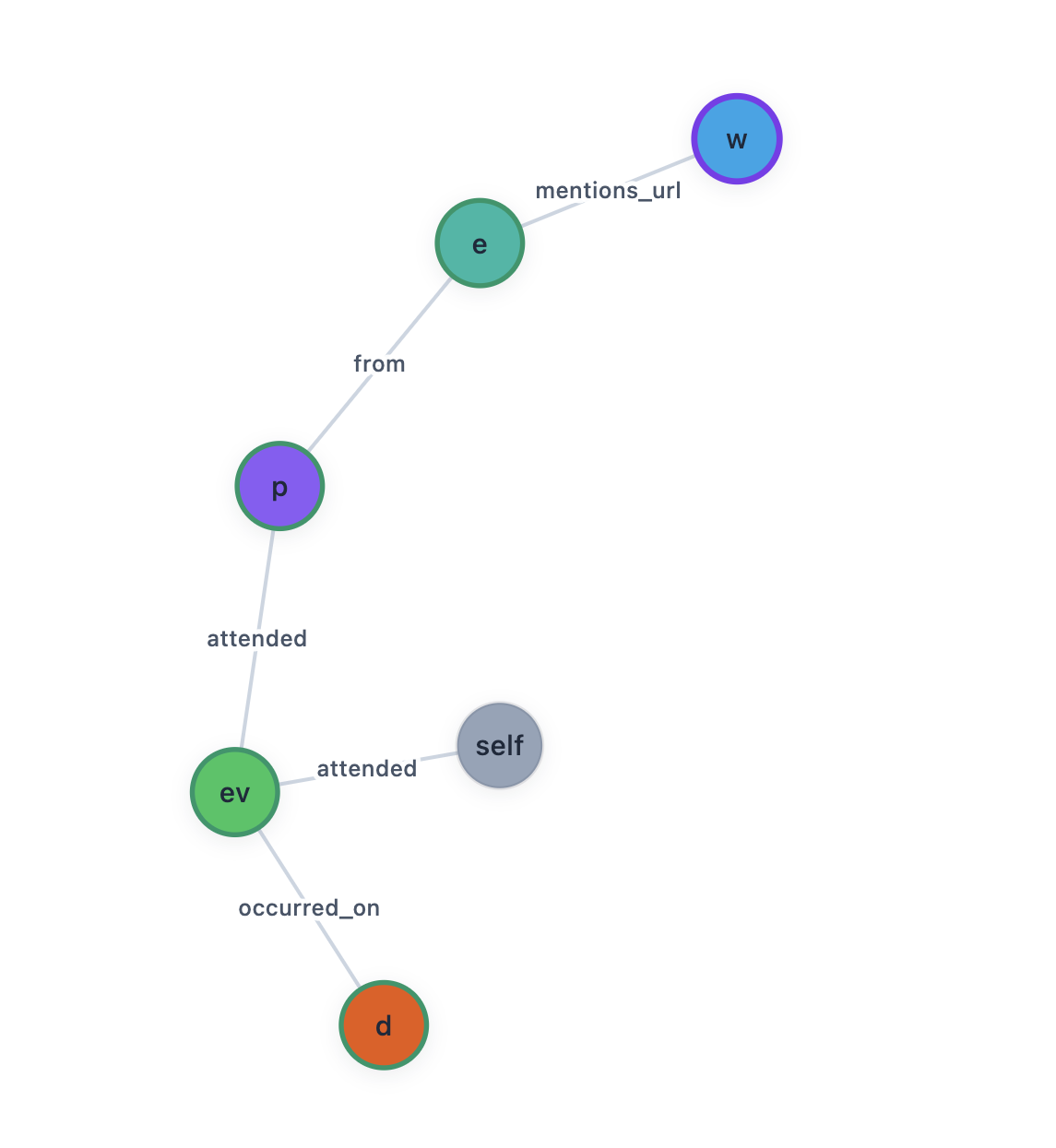}
    \textbf{Tool Execution Log} \\
     \begin{lstlisting}
    Next neighbors: [('e', 'mentions_url')]
    Next neighbors: [('p', 'from')]
    Next neighbors: [('ev', 'attended')]
    Next neighbors: [('d', 'occurred_on')]
    Next neighbors: []
    filtered date bound var size: 1
    tool to be called: date_occurred_on_event ev d
      result size: 57
    result size for ev: 57
    tool to be called: event_attended_person p ev
      result size: 4
    result size for p: 4
    tool to be called: person_from_email e p
      result size: 54
    result size for e: 54
    tool to be called: email_mentions_url_webpage w e
      result size: 159
    result size for w: 159
    Total query time (plan + tool run): 46.96s
    \end{lstlisting}
\end{enumerate}
...

\end{document}